\documentclass[letterpaper,journal,twoside]{IEEEtran}   
\usepackage{amssymb}  
\usepackage{graphicx}
\usepackage{stfloats}
\usepackage{amsmath}
\usepackage{gensymb}
\usepackage{subcaption}
\usepackage[table]{xcolor}
\usepackage{tabularx}
\usepackage{booktabs}
\usepackage{multirow}
\usepackage{cite}
\usepackage{float}
\usepackage{caption}
\usepackage{makecell}
\usepackage{siunitx}   
\usepackage{url}
\usepackage{etoolbox}

\title{\LARGE \bf
PUMA: Perception-driven Unified Foothold Prior for Mobility Augmented Quadruped Parkour
}
\author{Liang Wang$^{1,2}$, Kanzhong Yao$^{2}$, Yang Liu$^{2}$, Weikai Qin$^{2}$, Jun Wu$^{1}$, Zhe Sun$^{* 2}$ and Qiuguo Zhu$^{*1}$ 
\thanks{Manuscript received: January 22, 2026; Revised: May 7, 2026; Accepted: June 25, 2026.}%
\thanks{This paper was recommended for publication by Editor Olivier Stasse upon evaluation of the Associate Editor and Reviewers comments. This work was supported by the ``Leading Goose'' R\&D Program of Zhejiang (Grant No. 2023C01177), the National Key R\&D Program of China (Grant No. 2022YFB4701502), and the 2035 Key Technological Innovation Program of Ningbo City (Grant No. 2024Z300).}%
\thanks{$^{1}$The authors are with Institute of Cyber-Systems and Control, Zhejiang University, 310027, China
        \texttt{3210102182@zju.edu.cn}.}%
\thanks{$^{2}$Institute of Artificial Intelligence (TeleAI), China Telecom.}%
\thanks{$^{*}$Corresponding authors: Qiuguo Zhu (\texttt{qgzhu@zju.edu.cn}) and Zhe Sun (\texttt{sunzhe@nwpu.edu.cn}).}%
\thanks{Project website: \protect\url{https://puma-parkour.github.io/}.}%
\thanks{Digital Object Identifier (DOI): see top of this page.}%
}
\begin{document}
\markboth{IEEE ROBOTICS AND AUTOMATION LETTERS. PREPRINT VERSION. ACCEPTED JULY, 2026}%
{Wang \MakeLowercase{\textit{et al.}}: PUMA: Perception-driven Unified Foothold Prior for Mobility Augmented Quadruped Parkour}
\newcommand{\insertteaser}{
    \centering
    \includegraphics[width=0.98\linewidth,height=0.6\linewidth]{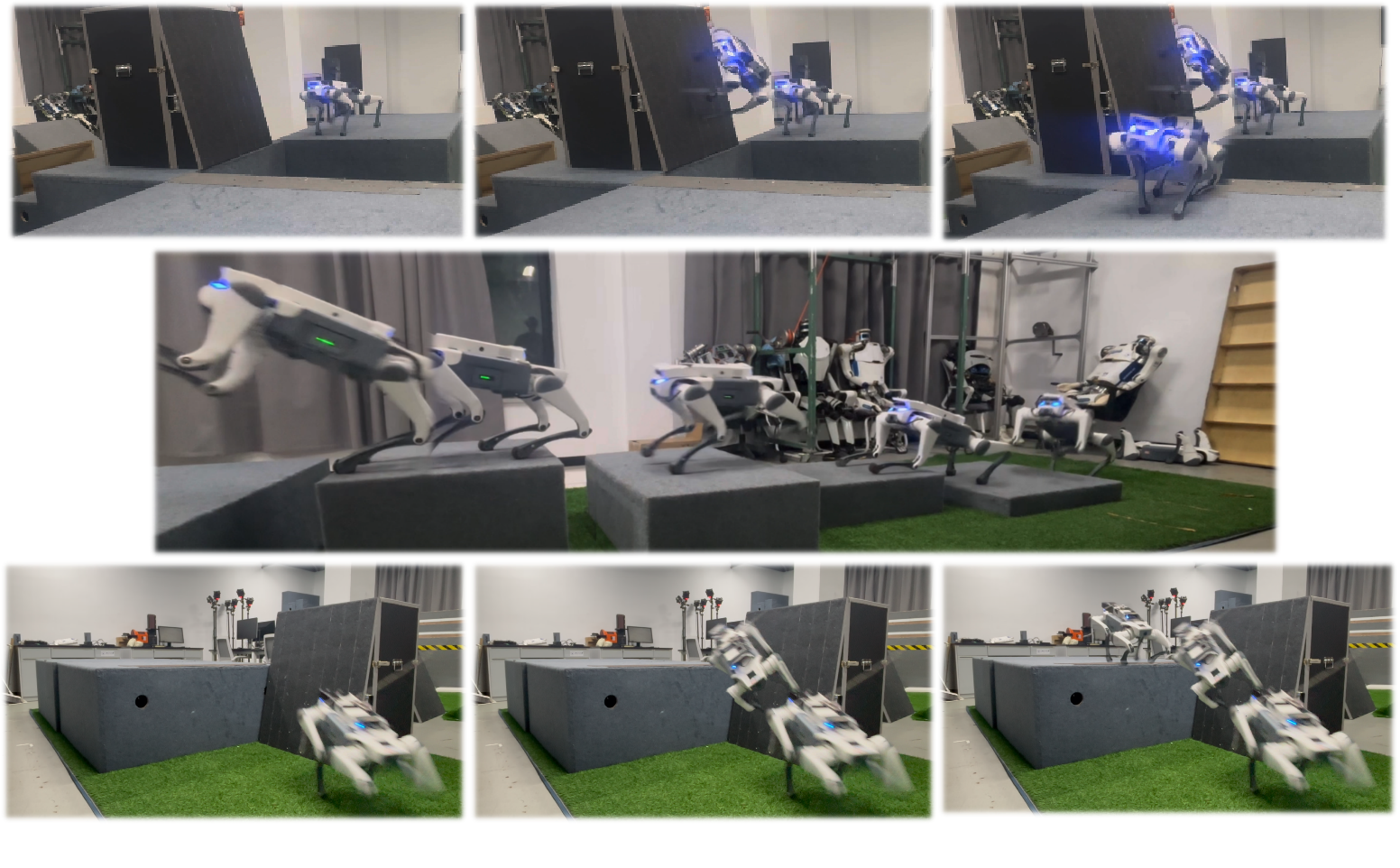}
    \vspace{-2em}
    \setcounter{figure}{0}
    \captionof{figure}{
        \textbf{PUMA} enables quadruped robots to fuse proprioception with visual perception to estimate adaptive footholds for traversing complex discrete terrains. 
        \textbf{Top Row:} The robot twists its posture to forcefully kick off the inclined wall, propelling itself across a wide gap.
        \textbf{Middle Row:} The robot sequentially traverses uneven stepping stones. 
        \textbf{Bottom Row:} The robot leverages an inclined wall to surmount a high platform.
    }
    \vspace{-3em}
    \label{fig:teaser}
}

\makeatletter
\apptocmd{\@maketitle}{\insertteaser}{}{}
\makeatother

\maketitle  
\setcounter{figure}{1}
\begin{abstract}
Parkour tasks for quadrupeds have emerged as a promising benchmark for agile locomotion. While human athletes can effectively perceive environmental characteristics to select appropriate footholds for obstacle traversal, endowing legged robots with similar perceptual reasoning remains a significant challenge. Existing methods often rely on hierarchical controllers that follow pre-computed footholds, thereby constraining the robot’s real-time adaptability and the exploratory potential of reinforcement learning. To overcome these challenges, we present PUMA, an end-to-end learning framework that integrates visual perception and foothold priors into a single-stage training process. This approach leverages terrain features to estimate egocentric polar foothold priors, composed of relative distance and heading, guiding the robot in active posture adaptation for parkour tasks. Extensive experiments conducted in simulation and real-world environments across various discrete complex terrains demonstrate PUMA's exceptional agility and robustness in challenging scenarios.  

\textit{Index Terms}: foothold prior, visual perception, reinforcement learning.
\end{abstract}
\section{INTRODUCTION}
In recent years, quadruped robots have demonstrated remarkable athletic performance: a low-cost quadruped robot can leap over long gaps, climb over high obstacles, and traverse complex discrete terrain composed of stepping stones and sloped platforms~\cite{kim2025high,luo2024pie,cheng2024extreme,zhuang2023robot,he2025attention,11303867}. To navigate complex terrains in parkour tasks, many studies leverage exteroceptive sensors (e.g., cameras, LiDAR) to acquire terrain information. The sensory data is typically fused with proprioception to train dynamic locomotion policies via learning-based methods. Approaches include both decoupled hierarchical frameworks separating control and perception modules~\cite{hoeller2024anymal,fu2022coupling}, and end-to-end visual-aided policies~\cite{yang2021learning,agarwal2023legged}, establishing a strong foundation for agile legged locomotion.

However, even with the integration of exteroceptive sensing, robots still struggle to fully comprehend or exploit terrain features. Human parkour athletes can leverage environmental features to extend locomotive potential beyond inherent physical limits, exemplified by kicking off a wall to gain extra height and reach otherwise inaccessible elevations. This ability to strategically leverage terrain features for accomplishing locomotion tasks beyond actuator constraints remains a critical challenge in current legged robots.  

To enable robots to traverse complex terrains with diverse geometric features, recent studies have concentrated on foot placement planning in locomotion. These approaches commonly adopt a hierarchical framework, where high-level trajectory optimization or learning-based methods are used to plan desired foothold locations based on terrain information. The planned footholds subsequently serve as execution objectives for the low-level tracking policy. Such methods have demonstrated robust locomotion capabilities across complex discrete terrains and deformable terrains~\cite{kim2025high,jenelten2024dtc,coholich2025hierarchical}. However, direct foothold-tracking methods rely on accurate and feasible foothold targets throughout execution. In highly dynamic parkour tasks, onboard perception can only observe a partial region of the environment, making it difficult to plan globally reliable footholds and obtain accurate target foothold locations at each control timestep. Therefore, strictly tracking the generated footholds may limit the policy's flexibility in adjusting its contacts and body motion during dynamic terrain interactions.

In this paper, we present PUMA, a \textbf{P}erception-driven \textbf{U}nified foothold prior framework for \textbf{M}obility \textbf{A}ugmented quadruped parkour. Unlike methods mentioned before, our method does not explicitly enforce foothold tracking. Instead, we employ egocentric polar footholds as motion priors for velocity tracking, which decomposes the explicit coordinates into relative distance and heading. By fusing depth perception with proprioception, PUMA estimates the polar footholds and feeds it directly into the actor network. To ensure stable convergence within this unified single-stage framework, we employ Probability Annealing Selection (PAS) method to gradually transition from ground-truth to predicted footholds during training. Extensive experiments on both simulated and physical robots equipped with onboard depth cameras demonstrate that PUMA enables robust traversal of challenging discrete terrains, such as uneven stepping stones, and allows the robot to clear wide gaps by strategically exploiting inclined walls, as shown in Fig.~\ref{fig:teaser}. The main contributions of this paper are as follows:  
\begin{itemize}
    \item A unified one-stage training framework that incorporates predicted footholds for robust locomotion over discrete terrains.
    \item A learned egocentric polar foothold prior that fuses proprioceptive and exteroceptive perception to guide velocity tracking without explicit foothold tracking.
    \item Extensive simulation and real-world experiments demonstrating robust sim-to-real transfer and agile locomotion using onboard depth sensing.
\end{itemize}

\section{RELATED WORK}

\subsection{Robot Parkour}

Quadruped robots have demonstrated agile motor skills with model-based methods \cite{nguyen2019optimized,nguyen2022continuous}. However, these approaches rely on precise environment modeling and exhibit limitations in complex, dynamically changing scenarios. In recent years, Reinforcement Learning (RL) has made significant progress in parkour tasks. Hoeller et al.~\cite{hoeller2024anymal} employed a hierarchical framework with separate locomotion control and perception modules, enabling navigation across highly complex terrains. Cheng et al.~\cite{cheng2024extreme} and Zhuang et al.~\cite{zhuang2023robot} utilized depth map information through a two-stage teacher-student structure to estimate privileged information and distill parkour locomotion policies. Luo et al.~\cite{luo2024pie} further integrated the dual-stage approach into a single-stage framework, leveraging an asymmetric actor-critic network to learn parkour policies capable of implicitly imagining privileged observations. Despite these advances, relying solely on RL exploration in the absence of prior knowledge limits the strategic exploitation of terrain features.

\subsection{Locomotion with Priors}

To enhance the efficiency of RL exploration in locomotion tasks, incorporating motion priors into learning frameworks has emerged as a key approach. A common paradigm is to leverage reference trajectories in conjunction with imitation learning, enabling quadruped robots to acquire agile and dynamic behaviors~\cite{peng2018deepmimic,peng2021amp}. These approaches, however, rely heavily on high-quality expert demonstrations, which are typically obtained through motion capture systems or manually curated datasets~\cite{escontrela2022adversarial,wu2023learning}. In practice, collecting such data is expensive and task-specific, and often requires additional processing or refinement, such as trajectory filtering, retargeting, or optimization-based smoothing~\cite{yoon2025spatio, 11128480}. This dependence limits scalability and reduces adaptability to new environments or tasks.

 \subsection{Foothold-based Locomotion}

 More recent approaches directly condition policy objectives on explicit footholds as task-specific targets. For instance, Kim et al.~\cite{kim2025high} leverage a hierarchical framework combining sampling-based foothold planning and a learning-based tracker module to achieve high-speed navigation over discrete terrains. Coholich et al.~\cite{coholich2025hierarchical} utilizes a high-level policy that optimizes foothold targets using the low-level policy's value function without additional training. Jenelten et al.~\cite{jenelten2024dtc} utilize trajectory optimization to generate optimized footholds combined with RL for robust tracking. These methods demonstrate that footholds can effectively enhance robot locomotion performance across complex terrains. However, explicit foothold tracking assumes accurate execution targets are available throughout traversal, which becomes demanding in highly dynamic parkour tasks with partial perception. To address this challenge, we introduce an egocentric foothold prior within a unified single-stage training framework, where locally estimated footholds are used as soft motion priors to achieve robust terrain traversal.

\section{METHOD}
\begin{figure*}[htbp]
    \centering
    \vspace{0.5em}
    \includegraphics[width=0.85\textwidth]{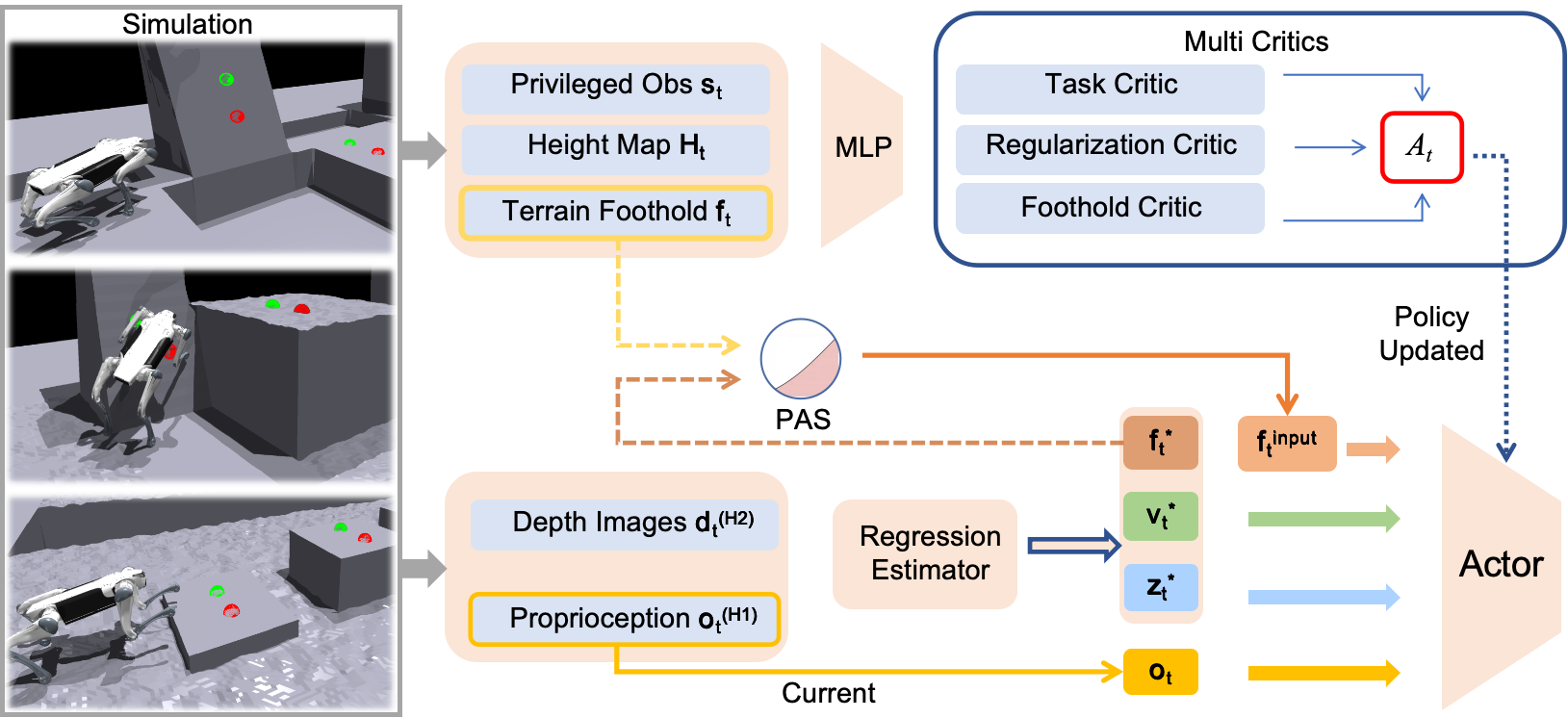}
    \caption{Overview of PUMA training framework. A velocity-tracking locomotion policy takes proprioception and depth images as input to predict egocentric foothold priors, base velocity, and latent terrain features. These representations are then concatenated with the current observations and fed into the policy network. Multiple critic networks are trained on distinct reward components to cooperatively optimize the policy. During training, a PAS strategy gradually replaces ground-truth footholds with predicted ones. The entire process is conducted in a single stage, with all networks optimized simultaneously.}
    \label{fig:pipeline}
    \vspace{-1.5em}
\end{figure*}

Our goal is to train an end-to-end velocity-tracking locomotion policy that integrates geometric features from an onboard depth camera with proprioception to accomplish parkour tasks over complex terrains. To address partial observability, we adopt an asymmetric actor-critic architecture~\cite{nahrendra2023dreamwaq,Pinto-RSS-18}, as illustrated in Fig.~\ref{fig:pipeline}. All modules in the training pipeline are optimized concurrently, trained from scratch without relying on pre-trained components.  

\vspace{-1em}
\subsection{Egocentric Foothold Prior Design}
In the onboard visual setting considered in this work, the policy only observes local terrain geometry, making it difficult to obtain globally reliable foothold plans. Therefore, the generated footholds are used to provide terrain-aware guidance for locomotion, rather than serving as precise tracking targets. 
 
We denote the sequence of candidate footholds as $\{\mathbf{p}_i\}_{i=0}^{N}$, where each $\mathbf{p}_i \in \mathbb{R}^3$ represents the Cartesian coordinate in the world frame. The sequence is first initialized from geometry-based anchors, such as the centers of sparse supports and traversable wall surfaces. When adjacent anchors are far apart, we further densify the sequence along the commanded direction with a 1\,m interval to keep the next candidate foothold within a local observable range. Since the footholds are used as guidance priors rather than explicit tracking targets, complex kinematic feasibility checks are unnecessary. Instead, we focus on local terrain validity and safety by filtering out candidates that fall outside valid traversable support regions or are situated too close to hazardous edges. The final valid foothold set is defined as:
\begin{equation}
\hspace*{-0.2cm}
\mathcal{P} = \left\{ \mathbf{p}_i \mid \mathbf{p}_i \in \mathcal{S}_{\mathrm{tr}},\  d_{\text{edge}}(\mathbf{p}_i) > d_{\text{safe}},\ i\in \{0, \dots, N\} \right\}
\end{equation}
where $\mathcal{S}_{\mathrm{tr}}$ denotes the locally valid traversable support region, $d_{\text{edge}}(\mathbf{p}_i)$ denotes the Euclidean distance from $\mathbf{p}_i$ to the nearest terrain edge, and $d_{\text{safe}}$ is the safety threshold.

To avoid overconstraining the learned gait adaptation, we focus exclusively on the front feet, which establish initial contact and dominantly shape the body attitude. In highly dynamic parkour motions, Cartesian foothold coordinates expressed in the robot frame can be sensitive to rapid body-attitude changes. We therefore adopt an egocentric polar representation to decouple relative distance from heading direction and provide a smoother foothold prior for policy learning. Formally, the feature vector $\mathbf{f}_t$ at time step $t$ is formulated as:
\[
\mathbf{f}_t = \left\{ d_t^{(L)},\ d_t^{(R)},\ \psi_t,\ \psi_{t+1} \right\}
\]

Here, $d_t^{(L)}$ and $d_t^{(R)}$ denote the Euclidean distances from the robot's left and right forefeet to the current expected foothold $\mathbf{p}_t$. $\psi_t$ and $\psi_{t+1}$ represent the heading error between the robot's current orientation and the target direction toward $\mathbf{p}_t$ and $\mathbf{p}_{t+1}$, respectively. 
\vspace{-0.5em}
\subsection{Training Pipeline}
Distinct from existing works~\cite{jenelten2024dtc,coholich2025hierarchical} that often adopt a hierarchical framework to separately handle foothold generation and locomotion, we employ a unified single-stage approach based on an asymmetric actor-critic architecture~\cite{nahrendra2023dreamwaq, Pinto-RSS-18}. Within this framework, the policy leverages the proposed egocentric foothold prior to directly learn parkour skills on complex terrains. The policy network is optimized using Proximal Policy Optimization (PPO).

\textbf{1) Network Inputs:} 
The policy network receives proprioceptive history $\mathbf{o}_{t}^{(H{1})}$ and depth image buffer $\mathbf{d}_{t}^{(H{2})}$ as input. Each proprioceptive observation $\mathbf{o}_{t}\in \mathbb{R}^{45}$ comprises the angular velocity $\boldsymbol{\omega}_t \in \mathbb{R}^{3}$, the gravity vector $\mathbf{g}_t \in \mathbb{R}^{3}$, the command $\mathbf{c}_t \in \mathbb{R}^{3}$, the joint positions $\boldsymbol{\theta}_t \in \mathbb{R}^{12}$, the joint velocities $\dot{\boldsymbol{\theta}_t} \in \mathbb{R}^{12}$, and the previous joint target positions $\mathbf{a}_{t-1} \in \mathbb{R}^{12}$.

The critic network, having access to privileged information, receives the privileged observation $\mathbf{s}_t$, surrounding heightfield $\mathbf {H}_t$ and the ground-truth egocentric foothold prior $\mathbf{f}_t$ as input. The privileged observation $\mathbf{s}_t$ is defined as:
\[
\mathbf{s}_t = \big[ \mathbf{o}_t,\ \mathbf{v}_t,\ \mathbf{p}_t,\ \mathbf{p}_{t+1}, \mathbf{x}_t \big]^T
\]
where $\mathbf{o}_t$ denotes the current proprioceptive observation, $\mathbf{v}_t \in \mathbb{R}^{3}$ is the base velocity. The terms $\mathbf{p}_t, \mathbf{p}_{t+1}$ represent the Cartesian coordinates of the current and next expected footholds, while $\mathbf{x}_t\in \mathbb{R}^{6}$ denotes the Cartesian positions of the robot's forefeet.

\textbf{2) Regression estimator:}
Building upon the implicit-explicit estimator proposed by Luo et al.~\cite{luo2024pie}, we extend this architecture to jointly infer the robot's internal states, environmental features, and the egocentric foothold prior from partial observations. 

Specifically, the depth image buffer $\mathbf{d}_{t}^{(H{2})}$ is first encoded via a Convolutional Neural Network (CNN). The extracted visual features are then concatenated with the proprioceptive history $\mathbf{o}_{t}^{(H{1})}$, forming a token sequence which is processed through a self-attention mechanism. The resulting sequence is fed into a Gated Recurrent Unit (GRU) network for temporal modeling, followed by several separate Multi-Layer Perceptron (MLP) heads to regress the estimated egocentric foothold prior $\hat{\mathbf{f}}_t$, the base velocity $\hat{\mathbf{v}}_t$, and the environment latent $\hat{\mathbf{z}}_t\in \mathbb{R}^{64}$.

\textbf{3) PAS in Foothold Prior:} As the actor network receives the estimated foothold $\hat{\mathbf{f}}_t$ from the regression estimator, unreliable priors in the early stage of training can destabilize the policy’s exploration, significantly impeding the learning process. To overcome the challenge, we apply the PAS method~\cite{11128428} to the foothold input, where the actor probabilistically receives either the ground-truth foothold or the estimator's prediction. Crucially, the probability of utilizing the ground-truth value gradually decreases as training progresses. At each training step, we sample a uniform random variable $u_t \sim \mathcal{U}(0,1)$ to decide whether to use the ground-truth foothold $\mathbf{f}_t$  or the estimator output $\hat{\mathbf{f}}_t$, based on the annealed probability $p_t$.
The annealing schedule is defined as:  
\begin{equation}
f_t^{\text{input}} = 
\begin{cases}
\hat{{f}}_t, & u_t < p_t, \\
{f}_t, & u_t \geq p_t,
\end{cases}
\quad u_t \sim \mathcal{U}(0,1)
\end{equation}
\begin{equation}
p_t = 1 - \cos\left(\frac{\pi t}{2T}\right)
\end{equation}
where $f_t^{\text{input}}$ denotes the input fed to the actor, $t$ is the current training iteration, and $T$ is the total number of annealing steps.

\subsection{Rewards and Multi Critics}
We design a composite reward structure based on the proposed egocentric polar foothold prior, incorporating heading alignment, dense distance tracking, and sparse arrival components. Beyond providing directional guidance, the reference footholds for the forefeet can implicitly encourage the robot to adapt its body posture to establish effective contact with the terrain. The foothold reward design is formulated as follows:
\begin{equation}
r_t^{\text{yaw}} = w_{\text{y}} \cdot\exp\left( - |\psi_t| \right)
\end{equation}
\begin{equation}
r_t^{\text{dense}} = w_{\text{d}} \cdot\exp\left( - \left( d_t^{(L)} + d_t^{(R)} \right) \right)
\end{equation}
\begin{equation}
r_t^{\text{sparse}} = w_{\text{s}} \cdot \mathbb{I}\left( d_t^{(L)} < \epsilon \land d_t^{(R)} < \epsilon \right)
\end{equation}

Here, \(w_y\), \(w_d\) and \(w_s\) denote the weights for yaw, dense and sparse rewards respectively. A fixed sparse reward is granted only when the distances of both the robot’s left and right forefeet to their corresponding expected footholds simultaneously fall within the threshold $\epsilon$.

\begin{table}[t]
    \centering
    \vspace{0.5em}
    \caption{Reward Functions and Weights}
    \label{tab:rewards}
    \renewcommand{\arraystretch}{0.2} 
    \begin{tabular}{@{}lcc@{}} 
        \toprule
        \textbf{Name} & \textbf{Equation} & \textbf{Weight} \\
        \midrule
        
        \textbf{Tracking Group} & & \textbf{3} \\
        Lin. vel tracking & $\exp\{-4\|\mathbf{v}_{xy}^{\text{cmd}} - \mathbf{v}_{xy}\|^2\}$ & $1$ \\
        Ang. vel tracking & $\exp\{-4\|\omega_{z}^{\text{cmd}} - \omega_{z}\|^2\}$ & $0.5$ \\
        \midrule
        
        \textbf{Foothold Group} & & \textbf{1.5} \\
        Dense Foothold & $\exp\left( - ( d_t^{(L)} + d_t^{(R)} ) \right)$ & $1$ \\
        Sparse Foothold &  $\mathbb{I}\left( d_t^{(L)} < \epsilon \land d_t^{(R)} < \epsilon \right)$ & $1$ \\
        Yaw Foothold    & $\exp\left( - |\psi_t| \right)$ & $1$ \\
        \midrule
        
        \textbf{Regularization Group} & & \textbf{1.0} \\
        Linear velocity ($z$) & $v_z^2$ & $-1.0$ \\
        Angular velocity ($xy$) & $\|\boldsymbol{\omega}_{xy}\|^2$ & $-0.05$ \\
        Orientation & $\|\mathbf{g}_{xy}\|^2$ & $-1.0$ \\
        Joint accelerations & $\|\ddot{\boldsymbol{\theta}}\|^2$ & $-2.5\text{e-}7$ \\
        Joint power & $\sum |\tau_j \dot{\theta}_j|$ & $-2.0\text{e-}5$ \\
        Collision & $n_{\text{col}}$ & $-10.0$ \\
        Action rate & $\|\mathbf{a}_t - \mathbf{a}_{t-1}\|^2$ & $-0.01$ \\
        Smoothness & $\|\mathbf{a}_t - 2\mathbf{a}_{t-1} + \mathbf{a}_{t-2}\|^2$ & $-0.01$ \\
        
        \bottomrule
    \end{tabular}
    \vspace{-1em}
\end{table}
To enhance value estimation under the mixed dense-sparse reward structure, we categorize the rewards into different groups based on their types and adopt Multi-Critic (MuC) approach~\cite{zargarbashi2025robotkeyframing,vijayan2025multi,mysore2022multi,HuangT-RSS-25} to independently estimate the return for each reward group:
\begin{equation}
r_t = w_{\text{task}} \cdot r_t^{\text{task}} + w_{\text{foothold}} \cdot r_t^{\text{foothold}} + w_{\text{reg}} \cdot r_t^{\text{reg}}
\end{equation}
Here, \(r_t^{\text{task}}\), \(r_t^{\text{foothold}}\), and \(r_t^{\text{reg}}\) represent the task, foothold, and regularization reward. The complete reward design is summarized in Table~\ref{tab:rewards}.  
Each critic network ${V}_{\phi_i}$ is optimized independently for its corresponding group with temporal difference loss:
\begin{equation}
L(\phi_i) = \hat{E}_{t} \left[ \left\| r_{i,t} + \gamma V_{\phi_i}(s_{t+1}) - \tilde{V}_{\phi_i}(s_{t}) \right\|^2 \right]
\end{equation}
Here, $i$ denotes the reward group index, $r_{i,t}$ is the scalar reward for group $i$ at step $t$, and $\tilde{V}_{\phi_i}$ represents the target value network. 

After each critic estimates the advantage associated with its corresponding reward group, the individual advantages are combined into a unified advantage used for policy optimization. Specifically, the weighted sum of advantages is normalized as:
\begin{equation}
\hat{A}_{\text{MuC}} = \frac{\sum_{i=0}^{n} w_i \cdot \hat{A}_i - \mu_{\text{MuC}}}{\sigma_{\text{MuC}}}
\end{equation}
\begin{equation}
L(\theta)=\mathbb{E}\bigl[\min\!\bigl(\alpha_t(\theta)\hat{A}_{\text{}},\,
\mathrm{clip}(\alpha_t(\theta),1\!-\!\epsilon,1\!+\!\epsilon)\,\hat{A}_{\text{}}\bigr)\bigr]
\end{equation}
where $\mu$ and $\sigma$ denote the batch mean and standard deviation of the weighted sum, $\alpha_t(\theta)$ is the probability ratio and $\epsilon$ the clipping hyper-parameter.

\subsection{Terrain and curriculum Design}
\label{subsec:terrain_chapter}
  
\begin{figure}[t]  
  \centering
  \includegraphics[width=1\linewidth]{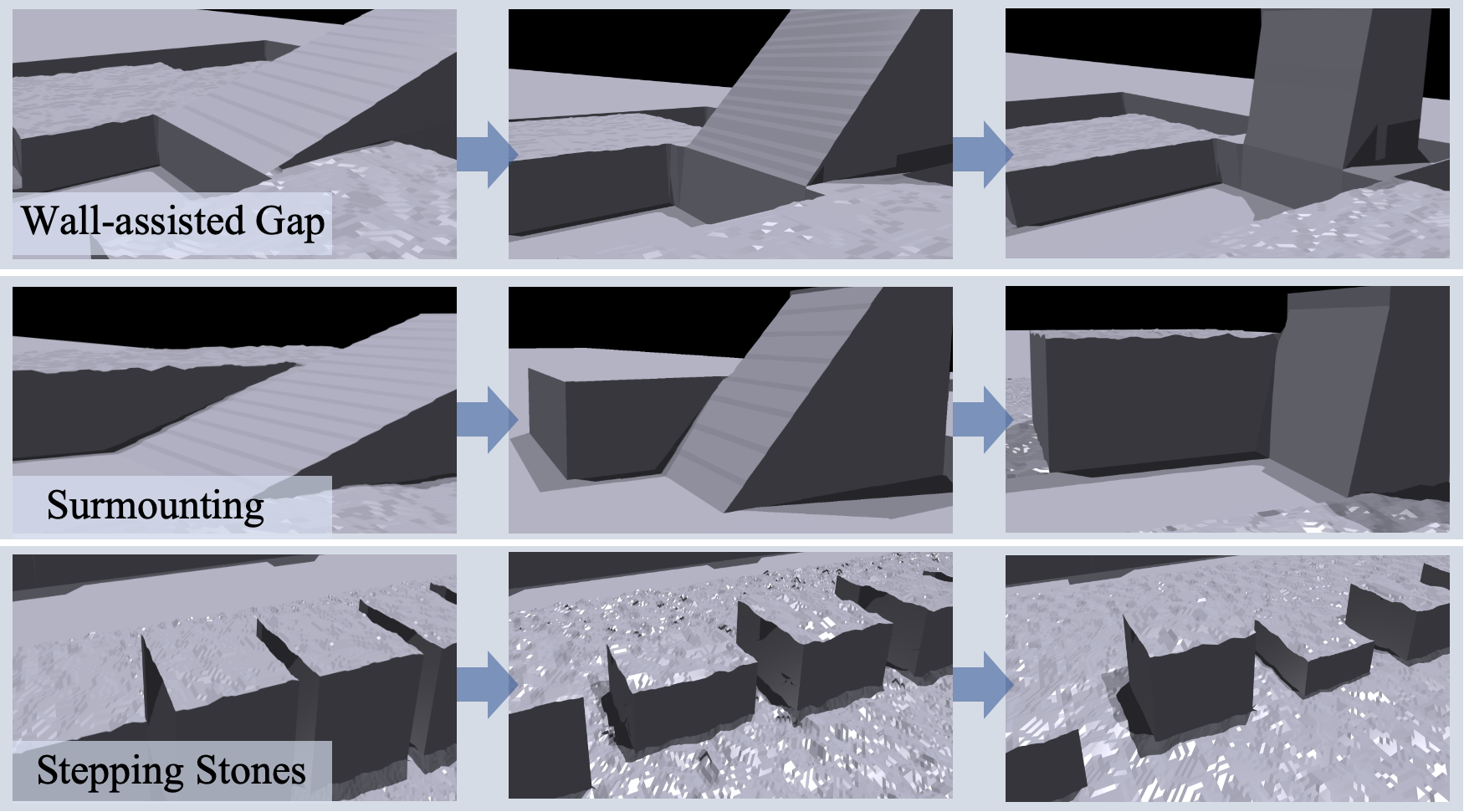}
  \caption{Terrain difficulty gradually increasing from the left side towards the right. Notably, we introduce height variations to horizontal surfaces to simulate roughness, while the inclined walls remain smooth.}
  \label{fig:terrain_curriculum}
  \vspace{-2em}
\end{figure}

As illustrated in Fig.~\ref{fig:terrain_curriculum}, we design three types of discrete terrains and establish a corresponding curriculum in simulation. 

\textbf{Wall-assisted Gap}: This terrain features randomly spaced gaps, each edged with inclined stepping walls. As the curriculum progresses, both the gap width and wall inclination increase. 

\textbf{Surmounting}: The terrain consists of elevated platforms with stepping walls attached to the leading edge. The platform height and wall inclinations are gradually increased throughout the curriculum. 

\textbf{Stepping Stones}: The terrain comprises stepping stones of varying dimensions and heights. As the curriculum level advances, the horizontal spacing and vertical drops between the stones increase, while the length and width decrease.

\section{EXPERIMENTS}

\begin{table*}[t]
\centering
\small
\setlength{\tabcolsep}{15pt} 
 \vspace{0.5em}
\caption{Success Rate (SR) and Traverse Rate (TR) comparison in simulation.
Each cell shows SR~TR (\%).}
\label{tab:results}
\begin{tabular}{l c | c c | c c}
\toprule
\multirow{2}{*}{Method} &
\multicolumn{1}{c}{Stepping Stone} &
\multicolumn{2}{c|}{Wall-assisted Gap} &
\multicolumn{2}{c}{Surmounting} \\
\multicolumn{1}{c}{} &
\multicolumn{1}{c}{SR~TR} &
\multicolumn{1}{c}{SR~TR} & \multicolumn{1}{c}{SR~TR} & 
\multicolumn{1}{c}{SR~TR} & \multicolumn{1}{c}{SR~TR} \\
\cmidrule(lr){2-2} \cmidrule(lr){3-4} \cmidrule(lr){5-6}
inclination angle & & 60\degree & 80\degree & 60\degree & 80\degree \\
\midrule
\textbf{Ours} & \textbf{98.7}~\textbf{99.4} & \textbf{98.6}~\textbf{97.3} & \textbf{96.2}~\textbf{95.3} & \textbf{96.8}~\textbf{96.9} & \textbf{94.7}~\textbf{92.5} \\
\midrule
\multicolumn{6}{l}{\textbf{(a) Ablation on foothold design}} \\
w/o Foothold Prior & 69.6~76.3 & 54.7~57.5 & 24.7~20.5 & 49.4~47.9 & 5.3~17.4 \\
w/o Relative Distance & 98.3~99.1 & 88.7~85.4 & 80.5~84.1 & 55.9~54.3 & 6.9~18.0 \\
Explicit Cartesian Prior & 91.6~90.1 & 93.5~91.9 & 86.4~85.2 & 90.9~88.3 & 78.1~81.9 \\
Implicit Cartesian Prior & 93.8~90.3 & 93.7~89.8 & 88.1~86.0 & 90.3~90.6 & 80.1~83.3 \\
\midrule
\multicolumn{6}{l}{\textbf{(b) Ablation on PAS iteration}} \\
w/o PAS & 98.1~98.3 & 96.2~96.6 & 93.1~94.5 & 94.8~97.3 & 93.2~94.5 \\

\midrule
\multicolumn{6}{l}{\textbf{(c) Ablation on critics}} \\
w/o MuC & 96.8~95.9 & 90.9~87.7 & 45.7~46.7 & 89.0~87.5 & 19.9~36.8 \\
\midrule
\multicolumn{6}{l}{\textbf{(d) Baselines}} \\
Explicit Foothold Tracker & 97.8~98.3 & 93.5~92.9 & 88.3~87.6 & 89.6~88.1 & 64.3~61.2 \\
PIE & 67.3~77.2 & 45.6~42.5 & 12.1~21.5 & 37.7~32.1 & --  ~13.9 \\
Extreme Parkour & 81.1~78.4 & 77.2~76.5 & 3.2~14.0 & 71.8~68.2 & --  ~14.3 \\
\bottomrule
\end{tabular}
\vspace{-1.5em}
\end{table*}

\subsection{Experimental Setup}

\textbf{Simulation training.} We train the 12-DoF DeepRobotics Lite3 robot using Isaac Gym across 2,048 environments on a single NVIDIA RTX 4090 GPU. The entire RL process is completed end-to-end without any pretraining. To maximize GPU memory efficiency during rollout, we implement a lightweight ray-tracing–based depth renderer built on NVIDIA Warp. To facilitate robust sim-to-real transfer, we further apply domain randomization on both observations and physical parameters, and introduce randomized delays in the visual input stream.  

\textbf{Real-world deployment.} The trained policy is deployed on Lite3 robot onboard using an RK3588 computing unit, running inference at 50\,Hz. The network outputs joint-level commands, which are translated into motor torques via PD control with gains $P=20$ and $D=0.5$. Depth observations are provided by the onboard Intel RealSense D435i camera, updating depth frames at 10\,Hz.   

We designed a set of experiments to demonstrate the effectiveness and advancement of our framework, as detailed below:

\textbf{(a)Ablation on foothold design}
\begin{itemize}
\item \textbf{w/o Foothold Prior:} The actor receives no foothold prior as input.
\item \textbf{w/o Relative Distance:} The foothold prior includes only the yaw angle component, omitting the distance estimates.
\item \textbf{Explicit Cartesian Prior:} The actor still uses footholds as motion priors, while the foothold prior is directly represented by Cartesian coordinates in the robot frame.
\item \textbf{Implicit Cartesian Prior:} The foothold prior is encoded as a compressed latent and reconstructed into Cartesian coordinates by an MLP decoder.
\end{itemize}

\textbf{(b)Ablation on PAS iteration}
\begin{itemize}
\item \textbf{w/o PAS:} Remove the PAS process during training.
\end{itemize}

\textbf{(c)Ablation on critics}
\begin{itemize}
\item \textbf{w/o MuC:} Employ a single critic network to estimate all reward functions.
\end{itemize}

\textbf{(d)Baselines}
\begin{itemize}
\item \textbf{Explicit Foothold Tracker:} A reference baseline for explicit foothold tracking. It directly receives explicit Cartesian foothold targets as policy input and uses foothold tracking as the primary objective without the velocity-tracking reward.
\item \textbf{PIE~\cite{luo2024pie}:} A single-stage parkour framework that enables the robot to traverse elevated platforms and gaps.
\item \textbf{Extreme Parkour~\cite{cheng2024extreme}:}  A two-stage parkour training framework that distills student policy from the teacher. We directly use the teacher policy here to bypass student limitations of imitation learning.
\end{itemize}

For PIE and Extreme Parkour, we retain their original reward functions while applying our terrain curriculum described in Section~\ref{subsec:terrain_chapter}.

\subsection{Simulation Experiments}
We conduct simulation experiments over the terrains described in Section~\ref{subsec:terrain_chapter} with a constant commanded robot velocity of 1.5 m/s. The locomotion performance of the different policies is measured using the following two metrics.
\begin{itemize}
  \item \textbf{Success Rate (SR):} The probability of successfully crossing the entire terrain segment.
  \item \textbf{Traverse Rate (TR):} The ratio of the furthest distance reached in an attempt to the total length of the terrain.
\end{itemize}

We present the success rates and traverse rates of all methods on the three terrain classes in Table~\ref{tab:results}, which are calculated from 1,000 trials in simulation. The experimental findings lead to the following insights:  

\textbf{1) Egocentric polar foothold prior extends the boundaries of parkour performance.}  

Our experimental results demonstrate that all methods incorporating foothold information outperform the baseline PIE and the ablation w/o foothold prior. PUMA, which employs the proposed egocentric polar foothold prior, achieves superior and robust performance across all terrain types. The w/o relative distance variant performs well on Stepping Stones, where yaw information is sufficient to maintain heading direction between discrete terrains, but degrades on Wall-assisted Gap and Surmounting, where the robot fails to learn the necessary body pose adjustment to make contact with highly inclined walls. The explicit and implicit Cartesian prior variants perform better than w/o relative distance on these terrains, since access to target foothold locations helps the robot adjust its body posture and achieve stable foot contact. However, they still underperform PUMA, especially on high-inclination terrains.

We additionally evaluate the accuracy of foothold regression on the Wall-assisted Gap and Surmounting terrains. As shown in Table~\ref{tab:foothold_accuracy}, PUMA, which predicts an egocentric polar prior, achieves significantly higher accuracy than the two Cartesian-based approaches. We attribute this superiority to the geometric nature of our representation: the polar format naturally decouples relative distance from heading direction and simplifies the regression landscape compared to coupled Cartesian coordinates $(x,y,z)$, enabling the network to learn a more precise and robust locomotion prior.

The Explicit Foothold Tracker performs well on Stepping Stone, but achieves lower SR/TR than PUMA on Wall-assisted Gap and Surmounting. We compare the provided footholds with the actual footholds taken by the robot. As shown in Table~\ref{tab:foothold_actual_error}, the Explicit Foothold Tracker achieves the smallest foothold errors. However, the footholds in our framework are generated by a simple terrain-heuristic strategy without strict dynamic feasibility verification. Therefore, strict tracking becomes more dependent on the quality and feasibility of the provided footholds, which can limit task performance in this setting. In contrast, PUMA allows the policy to adjust contacts within the tolerance range of the foothold reward, while using the posture and directional guidance from the foothold prior to traverse the terrain.
\begin{table}[htbp]
  \centering
  \small
  \setlength{\tabcolsep}{4pt}
  \caption{Foothold Regression MSE (\%).}
  \label{tab:foothold_accuracy}
  \begin{tabular}{l c c}
    \toprule
    Method & Wall-assisted Gap & Surmounting \\
    \midrule
    PUMA (Ours) & $6.33 \pm 0.73$ & $6.08 \pm 0.29$ \\
    Explicit Cartesian Prior & $12.10 \pm 1.20$ & $11.17 \pm 0.41$ \\
    Implicit Cartesian Prior & $9.30 \pm 0.98$ & $10.34 \pm 1.15$ \\
    \bottomrule
  \end{tabular}
  \vspace{-2em}
\end{table}  

\begin{table}[H]
\centering
\small
\setlength{\tabcolsep}{4pt}
\caption{Foothold error between provided and actual footholds in simulation (m).}
\label{tab:foothold_actual_error}
\begin{tabular}{l c c}
\toprule
Method & Wall-assisted Gap & Surmounting \\
\midrule
PUMA (Ours) & $0.0930 \pm 0.0417$ & $0.1210 \pm 0.0657$ \\
Explicit Cartesian Prior & $0.1383 \pm 0.0875$ & $0.1778 \pm 0.0804$ \\
Implicit Cartesian Prior & $0.1456 \pm 0.0733$ & $0.1565 \pm 0.0726$ \\
Explicit Foothold Tracker & $0.0695 \pm 0.0537$ & $0.0774 \pm 0.0685$ \\
\bottomrule
\end{tabular}
\vspace{-1em}
\end{table}

\begin{figure}[t]
    \centering
     \vspace{0.5em}
    \includegraphics[width=0.48\textwidth]{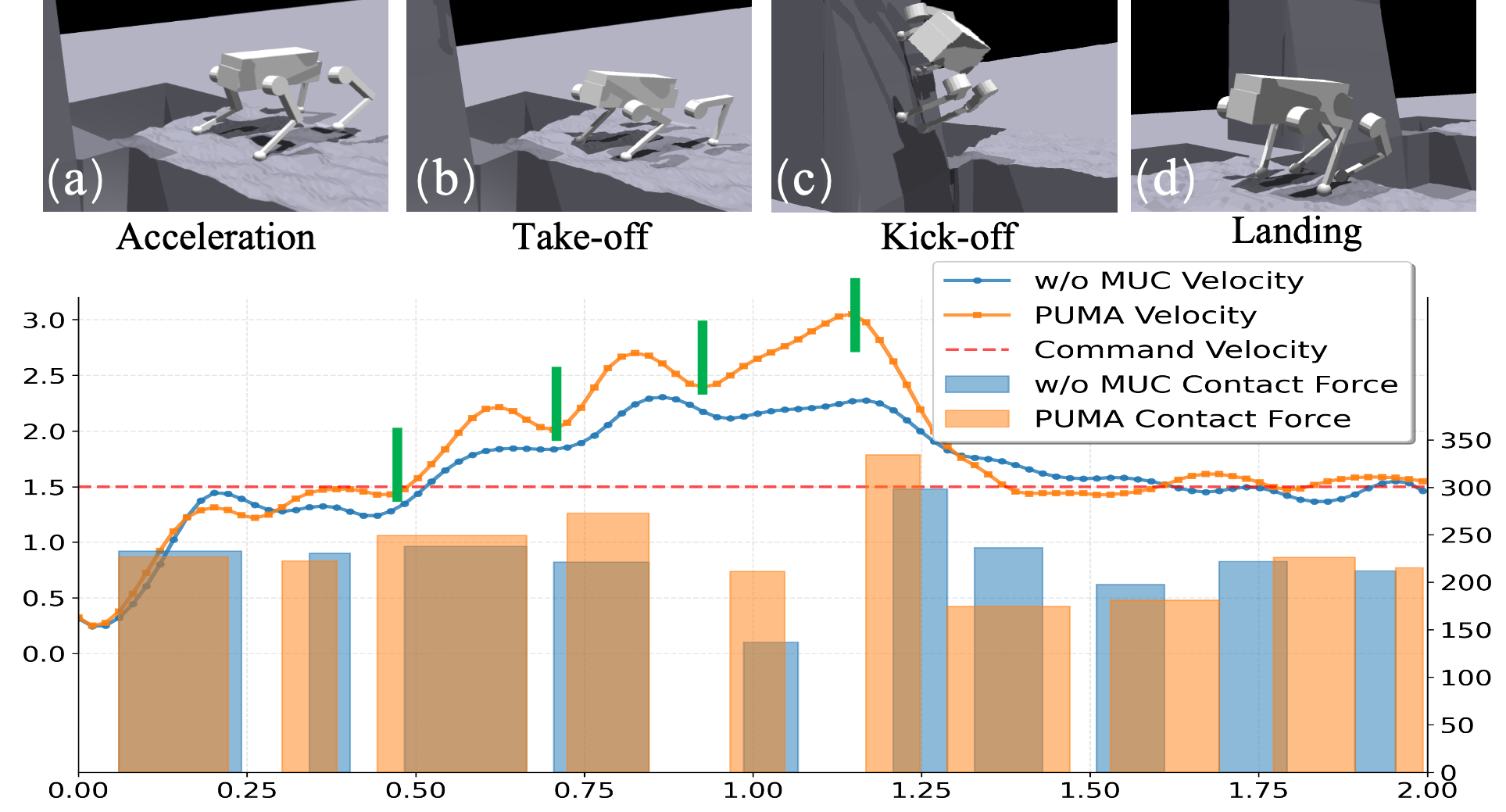} 
    \caption{Temporal evolution of body velocity and total contact force during a complete jump motion. The four motion phases are visualized in the top row and delimited by vertical green markers. The curves compare the body velocity of PUMA (orange) and the w/o MuC (blue) against the reference velocity (red dashed line). The bottom bars denote continuous intervals with nonzero contact force: the bar width indicates the interval duration, and the bar height indicates the average total contact-force magnitude over the four feet within that interval.}
    \label{fig:comparison_velocity_force_combined}
    \vspace{-1.5em}
\end{figure}

\textbf{2) The MuC design improves value estimation under mixed dense-sparse rewards.} 

\begin{figure}[t]
  \centering
  \begin{subfigure}[t]{0.49\linewidth} 
    \centering
    \includegraphics[width=\linewidth]{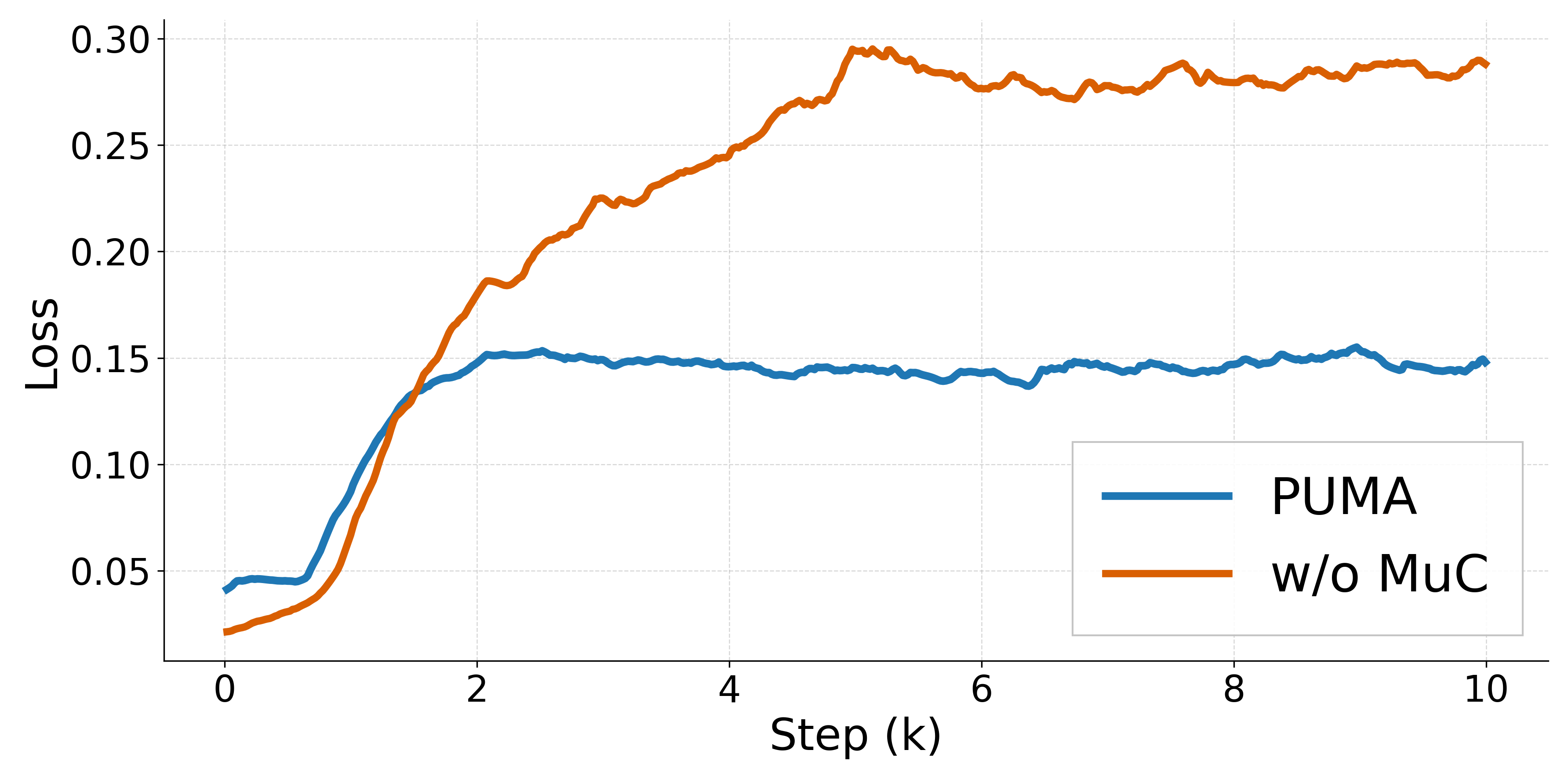}
    \caption{Value Loss}
    \label{fig:value_loss}
  \end{subfigure}
  \hfill
  \begin{subfigure}[t]{0.49\linewidth}
    \centering
    \includegraphics[width=\linewidth]{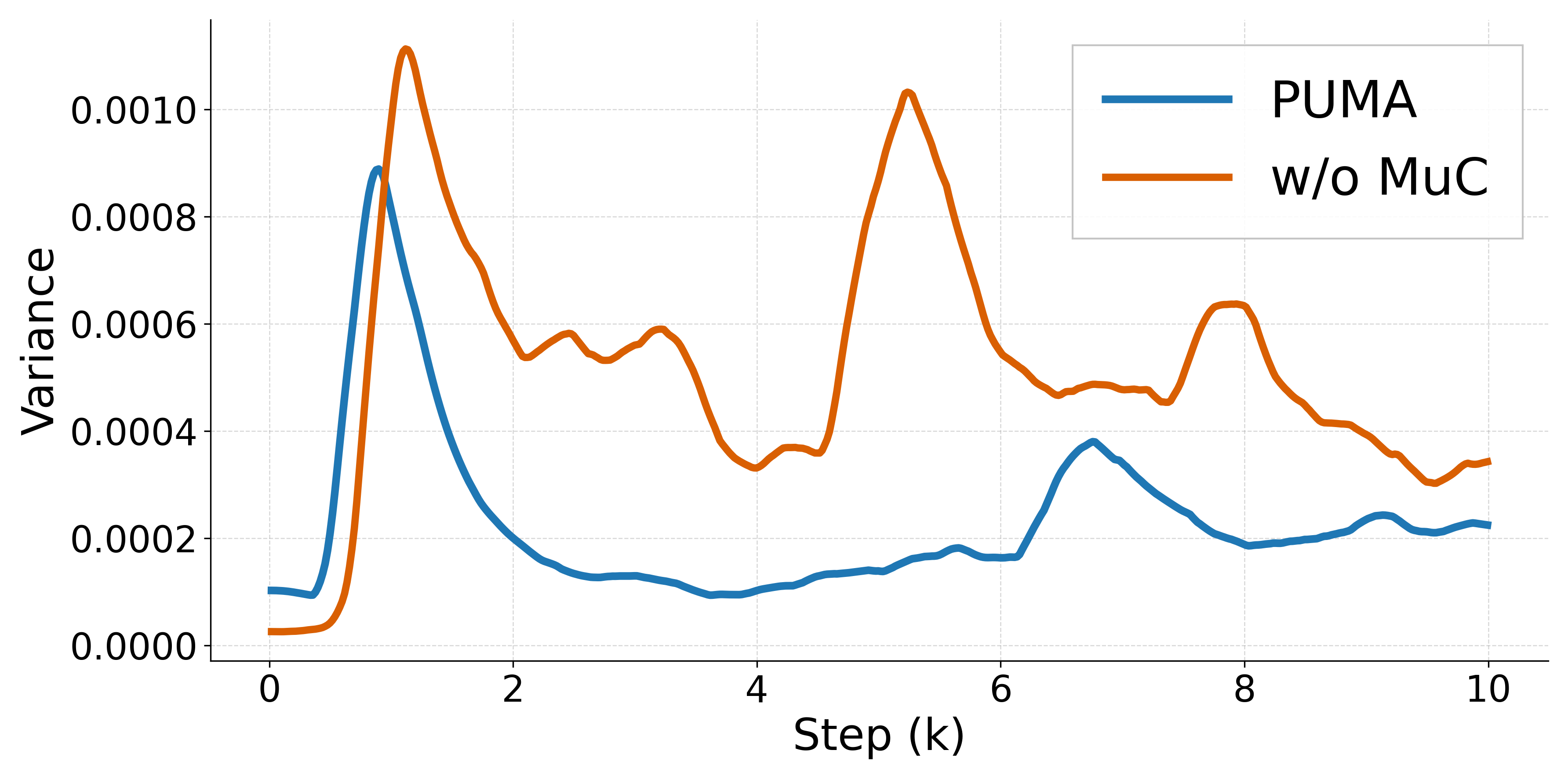}
    \caption{Value Loss Variance}
    \label{fig:value_loss_variance}
  \end{subfigure}

  \vspace{0.35em}

  \begin{subfigure}[t]{0.49\linewidth} 
    \centering
    \includegraphics[width=\linewidth]{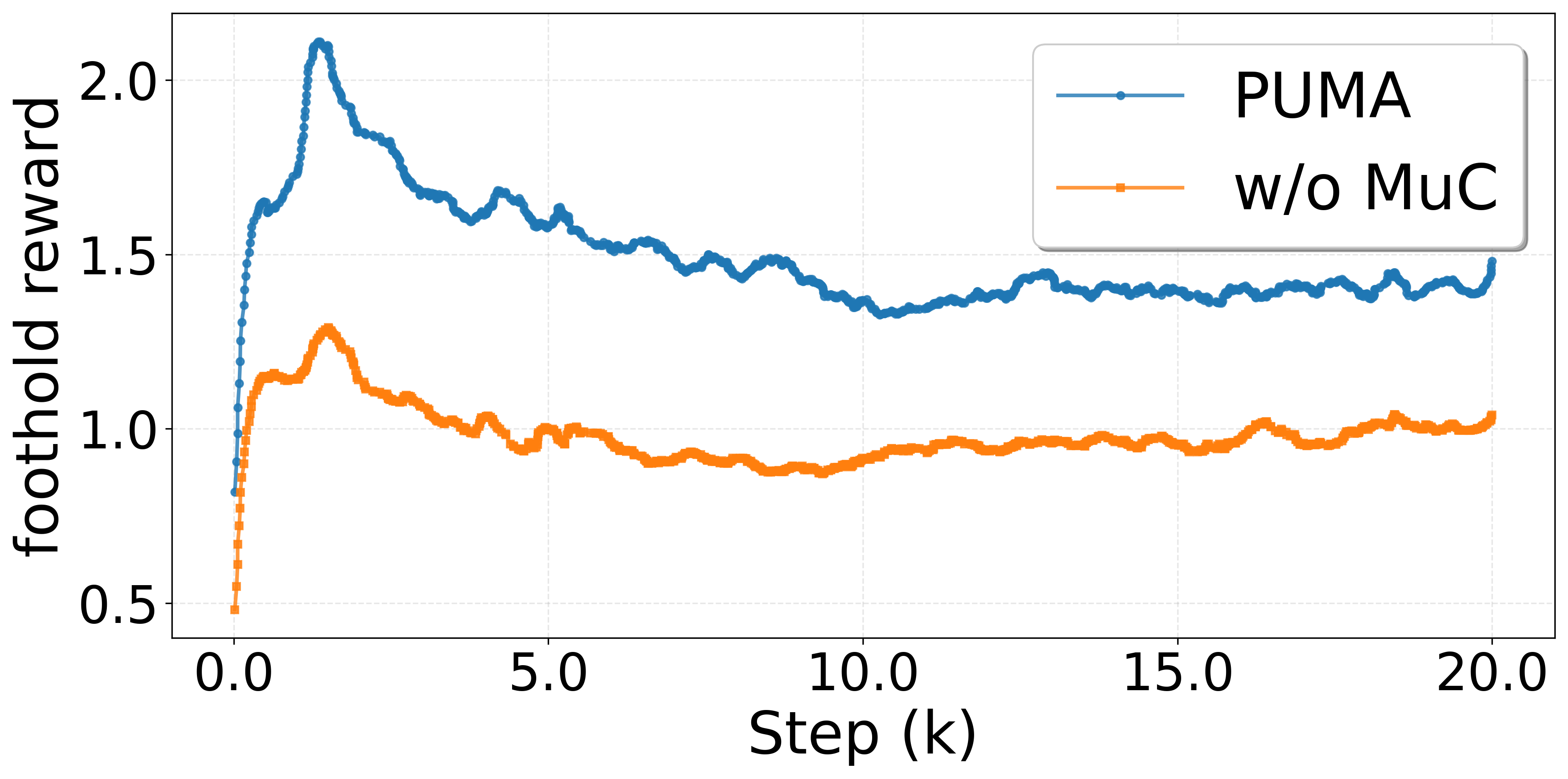}
    \caption{Foothold Rewards} 
    \label{fig:rewards} 
  \end{subfigure}
  \hfill
  \begin{subfigure}[t]{0.49\linewidth}
    \centering
    \includegraphics[width=\linewidth]{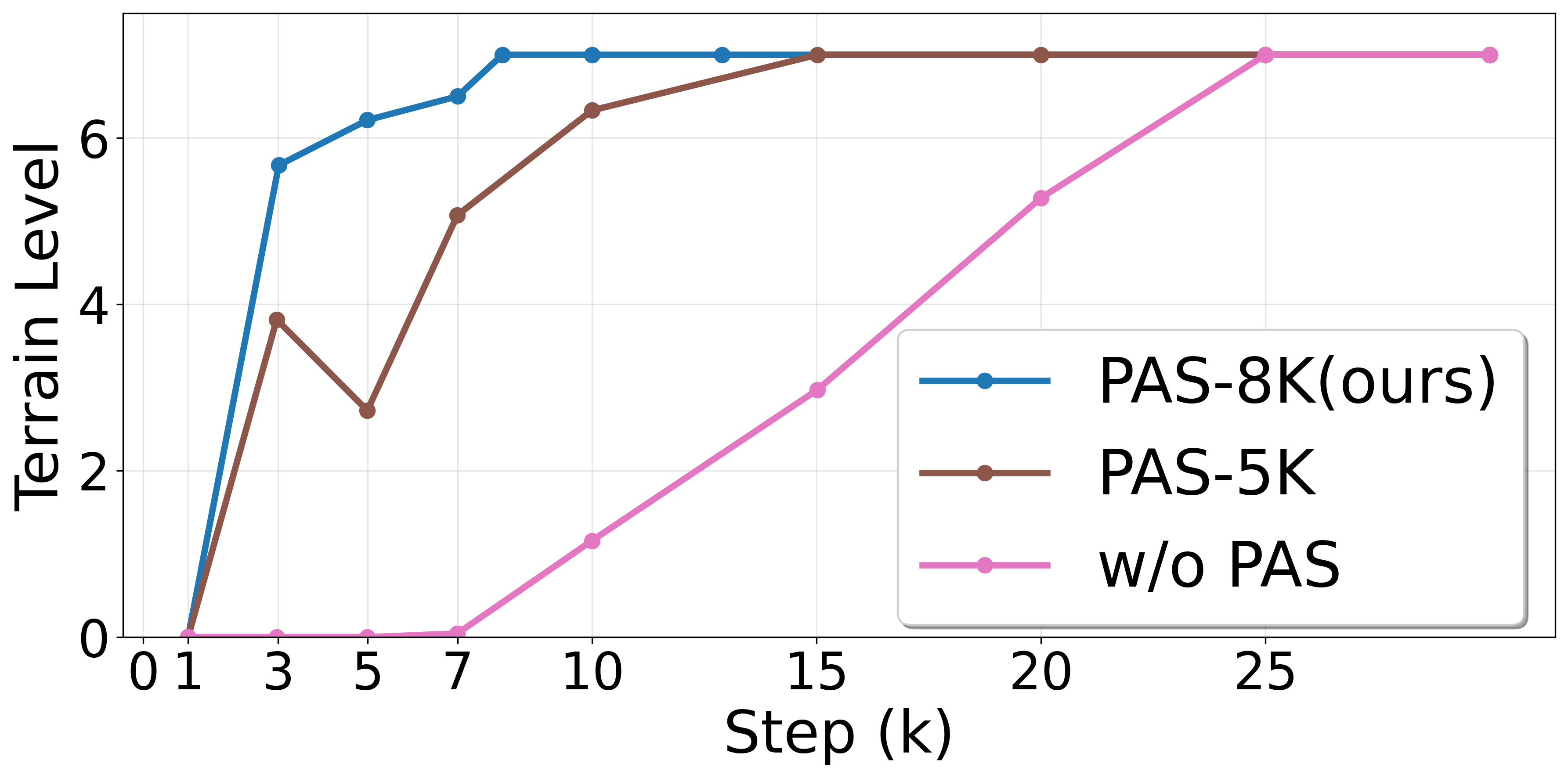}
    \caption{Learning Efficiency}
    \label{fig:efficiency}
  \end{subfigure}

  \caption{Training curve analysis. (a) Value loss. (b) Variance of value loss. (c) Foothold rewards. (d) Learning efficiency under different PAS annealing schedules, where 5K and 8K denote the annealing completion steps.}

  \label{fig:combined_analysis}
  \vspace{-2em}
\end{figure}

The success rate of w/o MuC drops significantly on highly inclined terrains. We analyze the robot's velocity and contact forces during wall-assisted jumps, as shown in Fig.~\ref{fig:comparison_velocity_force_combined}. PUMA exerts greater forces during the take-off and kick-off phases and reaches higher peak velocity, allowing the robot to exploit the reaction force from the inclined wall. In contrast, w/o MuC generates insufficient thrust during these contact phases and fails to sustain the subsequent motion, leading to task failure. We further evaluate the accuracy and stability of value estimation. As shown in Fig.~\ref{fig:value_loss} and Fig.~\ref{fig:value_loss_variance}, MuC achieves lower value loss and smaller variance during training, indicating more stable value estimation under the mixed dense-sparse reward structure. This is consistent with the reward curves in Fig.~\ref{fig:rewards}, where w/o MuC fails to effectively acquire the foothold reward. These results suggest that MuC helps the policy acquire foothold rewards more effectively through improved value estimation.

\begin{figure*}[htbp]
    \centering
    \vspace{0.5em}
    \includegraphics[scale=0.63]{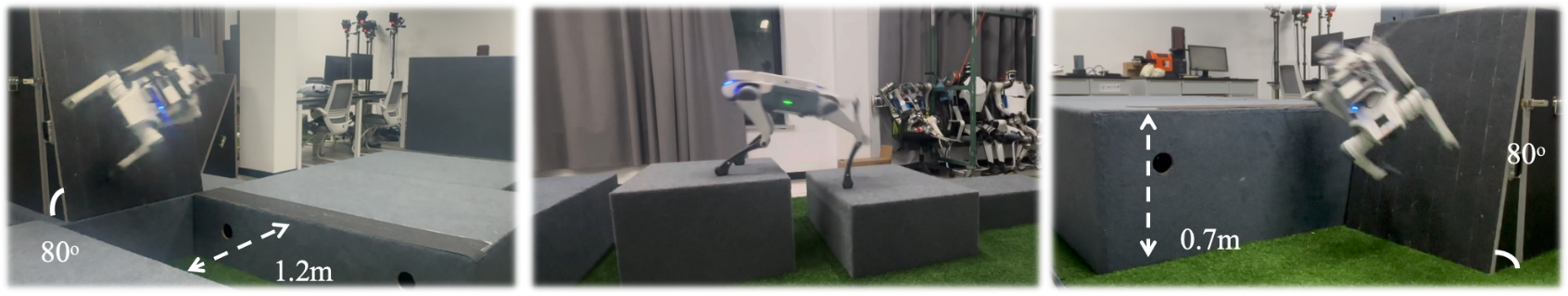}
    \vspace{-1em}
    
    \setlength{\tabcolsep}{2pt}

    \begin{minipage}[t]{0.32\textwidth} 
        \centering
        
        \resizebox{\linewidth}{!}{
        \begin{tabular}{lcc}
            \toprule
            & \multicolumn{2}{c}{\textbf{Success Rate}} \\
            \cmidrule(lr){2-3}
            \textbf{Method} & \textbf{60°} & \textbf{80°} \\
            \midrule
            \cellcolor{gray!10}PUMA (Ours) & \cellcolor{gray!10}1.0 & \cellcolor{gray!10}1.0 \\
            w/o Relative Distance & 0.2 & 0.0 \\
            Explicit Cartesian Prior & 0.6 & 0.2 \\
            Implicit Cartesian Prior & 0.7 & 0.3 \\
            w/o MuC & 0.6 & 0.0 \\
            \bottomrule
        \end{tabular}
        }
    \end{minipage}
    \hfill
    \begin{minipage}[t]{0.32\textwidth}
        \centering
        \resizebox{\linewidth}{!}{
        \begin{tabular}{lc}
            \toprule
            \textbf{Method} & \textbf{Success Rate} \\
            \midrule
            \cellcolor{gray!10}PUMA (Ours) & \cellcolor{gray!10}1.0 \\
            \cellcolor{gray!10}w/o Relative Distance & \cellcolor{gray!10}1.0 \\
            Explicit Cartesian Prior & 0.5 \\
            Implicit Cartesian Prior & 0.6 \\
            w/o MuC & 0.7 \\
            \bottomrule
        \end{tabular}
        }
    \end{minipage}
    \hfill
    \begin{minipage}[t]{0.32\textwidth}
        \centering
        \resizebox{\linewidth}{!}{
        \begin{tabular}{lcc}
            \toprule
            & \multicolumn{2}{c}{\textbf{Success Rate}} \\
            \cmidrule(lr){2-3}
            \textbf{Method} & \textbf{60°} & \textbf{80°} \\
            \midrule
            \cellcolor{gray!10}PUMA (Ours) & \cellcolor{gray!10}1.0 & \cellcolor{gray!10}0.8 \\
            w/o Relative Distance & 0.0 & 0.0 \\
            Explicit Cartesian Prior & 0.4 & 0.1 \\
            Implicit Cartesian Prior & 0.4 & 0.1 \\
            w/o MuC & 0.2 & 0.0 \\
            \bottomrule
        \end{tabular}
        }
    \end{minipage}
    
    \caption{Real-world experimental results: (left) Wall-assisted Gap terrain, featuring stepping walls at 60° and 80° angles followed by a 1.2\,m wide gap; (center) Stepping Stones terrain composed of discrete stones 0.5--0.8\,m in length/width and 0--0.4\,m in height variation; (right) Surmounting terrain with stepping walls at 60° and 80° angles leading to a 0.7\,m high platform.}
    \label{fig:real_world_results}
    \vspace{-1em}
\end{figure*}

\textbf{3) PAS boosts learning in concurrent network training.} 

Our results indicate that the PAS method does not correlate with the locomotion performance. Through an additional ablation study on the annealing duration, we find that the efficacy of PAS is primarily manifested in a marked acceleration of policy convergence during training. The learning efficiency across different methods is illustrated in Fig.~\ref{fig:efficiency}. The training of the w/o PAS method progressed at a markedly slow pace, hardly learning an effective policy in the early stage. In comparison, the PAS-5K method and our approach exhibited comparable learning efficiency. However, the performance of PAS-5K degrades significantly during the late annealing phase. This decline likely stems from the premature conclusion of the annealing schedule, which thereby forces the actor to rely on the regression estimator before it has fully converged. The resulting inaccurate foothold priors act as noisy inputs, destabilizing the policy and leading to a decline in the curriculum level.
\vspace{-1em}
\subsection{Real World Experiments}
\begin{figure}[t]
    \centering
    \includegraphics[width=1\linewidth]{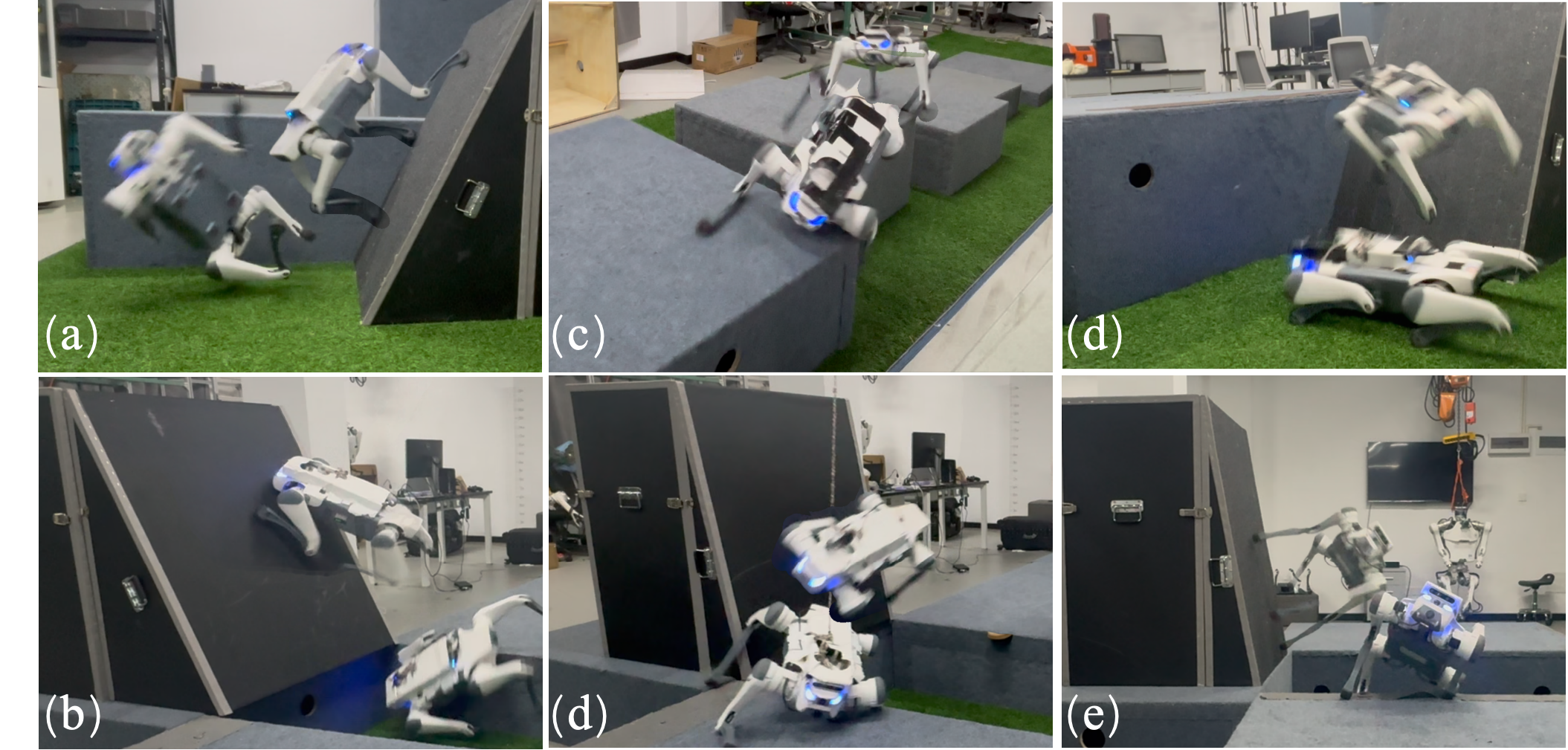}
    \caption{Typical failure cases in the real-world experiment: 
(a)--(b) Correct yaw but failed roll adaptation leading to sliding down (a) or collision (b); 
(c) Misorientation leading to a missed step; 
(d) Foothold estimation error causing missed wall contact; 
(e) Insufficient jump height causing platform collision; 
(f) The robot modifies its body posture to make contact with the inclined wall but fails to generate effective propulsion, resulting in an insufficient jump distance.}
\label{fig:fail_case}
    \vspace{-2em}
\end{figure}
 We evaluate the zero-shot transfer performance of our proposed method and the ablations in real-world scenarios, as shown in Fig.~\ref{fig:real_world_results}. The evaluation is conducted across each terrain with 10 trials performed for each method.

The physical experiments demonstrate that our approach enables the robot to execute highly dynamic parkour tasks. Notably, the robot autonomously exhibits a galloping gait and can rapidly traverse discrete terrains. It effectively leverages stepping walls to gain kinetic energy, allowing it to leap across wide gaps and vault over high platforms.

Additionally, we analyze the failure cases encountered during the real-world experiments, with the results presented in Fig.~\ref{fig:fail_case}. Due to multifaceted noise in the real-world environment, the foothold estimation accuracy of the Explicit Cartesian Prior and Implicit Cartesian Prior in foothold methods remains unstable, often leading to erroneous take-offs. These errors manifest primarily in incorrect body orientation and improper relative height during wall contact. While the w/o Relative Distance lacks the capability to adjust body posture for leveraging terrain features, it is noteworthy that it performs adequately on stepping-stone terrain. The w/o MuC method demonstrates poor real-world performance: although it can modify body posture, it consistently fails to establish effective force contact with the stepping wall, often resulting in false or grazing contact. In contrast, our proposed method successfully mitigates these issues, exhibiting remarkable stability and robustness.  

\section{CONCLUSION}

In this work, we proposed PUMA, an end-to-end learning framework that enables quadruped robots to perceive terrain geometry from onboard visual sensing, infer egocentric foothold priors, and leverage them as motion guidance for robust locomotion over discrete and challenging terrains. By integrating foothold-aware perception with a unified learning framework, PUMA enables stable posture adaptation and dynamic traversal without relying on explicit foothold tracking or hierarchical planning.

Despite these promising results, several limitations remain. The current framework does not address automatic multi-path foothold planning and remains limited on terrains where the left and right foothold regions are staggered. Future work will extend the framework toward perception-based multi-path foothold planning and broader sparse-terrain settings to improve behavioral diversity and generalization ability in more complex environments.

\addtolength{\textheight}{-12cm} 

\bibliographystyle{IEEEtran}   
\bibliography{reference}     

\end{document}